\pdfoutput=1
\documentclass[a4paper,twocolumn,hidelinks]{article}
\usepackage{isairas_arxiv_v1}

\begin{document}
\title{Overcoming the Challenges of Solar Rover Autonomy: Enabling Long-Duration Planetary Navigation }
\author[1]{Olivier Lamarre}
\author[2]{Jonathan Kelly}
\affil[1]{STARS Laboratory, University of Toronto, Canada, E-mail: \href{mailto:olivier.lamarre@robotics.utias.utoronto.ca}{olivier.lamarre@robotics.utias.utoronto.ca} }
\affil[2]{STARS Laboratory, University of Toronto, Canada, E-mail: \href{mailto:jkelly@utias.utoronto.ca}{jkelly@utias.utoronto.ca} }

\maketitle
\thispagestyle{empty}

\section*{Abstract}

The successes of previous and current Mars rovers have encouraged space agencies worldwide to pursue additional planetary exploration missions with more ambitious navigation goals. For example, NASA's planned Mars Sample Return mission will be a multi-year undertaking that will require a solar-powered rover to drive over 150 metres per sol for approximately three months. This paper reviews the mobility planning framework used by current rovers and surveys the major challenges involved in continuous long-distance navigation on the Red Planet. It also discusses recent work related to environment-aware and energy-aware navigation, and provides a perspective on how such work may eventually allow a solar-powered rover to achieve autonomous long-distance navigation on Mars.

\section{Introduction}

Surface mobility on Mars has tremendously accelerated the study of our solar system neighbour. In 1997, the National Aeronautics and Space Administration (NASA) successfully delivered the Sojourner rover to the surface of the Red Planet, as part of the Mars Pathfinder mission. This little rover was the first robot to navigate on another planet. Several years later, in 2004, the successful landings of the Mars Exploration Rovers (MERs), Spirit and Opportunity, marked the beginning of an active and ongoing mobile presence on Mars, which continues to the present day. In 2012, the Curiosity rover, part of the Mars Science Laboratory (MSL) mission, became the largest rover to date to explore the Martian surface.

Despite the very large amount of data collected by rovers across different regions of Mars, the capability of these rovers to conduct experiments on soil samples, for example, is very limited. This is why NASA is preparing a three-phase endeavour, called the Mars Sample Return (MSR) mission, to return Martian soil samples to Earth. The first phase of the mission will involve using the Mars 2020 rover to cache samples and drop them at a prespecified location \cite{mars2020}. The second phase will then land both a solar-powered rover and an ascent vehicle on the surface, to fetch the cache container and launch it into Martian orbit, respectively. As part of the third and final phase, an orbiter will capture the sample container and return it to Earth. 

In 2011, it was reported that the ``fetch'' rover for the MSR mission would be expected to drive up to fourteen kilometres in approximately three months \cite{mattingly2011mars}, corresponding to more than 150 metres per sol. Technically, Opportunity and Curiosity are already capable of driving such a distance during a single sol, but, in reality, this pace cannot be maintained. In fact, since the beginning of its mission, Curiosity has driven over 130 metres during a single sol only four times \cite{mslnaiflog}. This can be attributed to numerous factors, including operational restrictions, flight hardware limitations, and constraints related to the mission's science-driven goals. Nevertheless, such limitations raise questions about the ability of future solar-powered mobile robots, like the MSR fetch rover, to navigate long distances at a relatively fast pace across the diverse Martian terrain.

Numerous survey and review papers have already discussed the future of mobile exploration of Mars (e.g., the authors of \cite{wong2017adaptive} review generic adaptive capabilities and behaviours of rovers operating within a fully autonomous framework). However, to the best of our knowledge, no paper has described the challenges of rover navigation in energy-limited settings, considering the present operational context and near-future mission goals. This paper surveys how the current operational and technological challenges of planetary mobility planning and execution can be addressed by recent research in the field. More specifically, we identify work that could be applied in the context of long-distance navigation autonomy with solar-powered rovers, to meet the navigation requirements of the MSR mission and beyond.

The remainder of the paper is structured as follows: Section \ref{current} describes the state of the art in navigation planning for Mars rovers, while Section \ref{challenges} highlights the current challenges related to mobility on Mars. Section \ref{work} introduces promising work towards energy-aware navigation; Section \ref{discussion} discusses how the per-sol driving range of solar-powered rovers may be substantially improved over the next few years.

\section{Current Rover Navigation Framework}
\label{current}

Navigating on Mars is very challenging. In order to understand the implications of the research presented in the remainder of this paper, it is necessary to be familiar with the current rover navigation planning and execution process. This section provides a very brief description of the operational framework used to drive Curiosity and outlines the rover's navigation capabilities.

\subsection{Navigation Planning}

The operational framework employed to control Curiosity is composed of three distinct cycles: strategic, supratactical, and tactical. The strategic cycle can vary from a few weeks to a few months, and involves high-level and long-term activity planning; it heavily relies on input from the science teams to select sites to investigate (based primarily on orbital data), incorporates preliminary traversability and activity assessment, and continuously ensures that the mission maintains focus on its objectives while accounting for changes in the rover's condition and capabilities \cite{chattopadhyay2014mars}. The tactical cycle, on the other hand, lies at the opposite end of the planning spectrum, typically lasting a day (or a few days at most). The purpose of this cycle is to ensure that daily mission goals, formulated by the strategic and supratactical teams, are fulfilled. Tactical mission planners mainly utilize surface data collected by Curiosity's sensors (such as stereo imagery and telemetry). Tactical planning is highly reactive: on every cycle, the tactical planners analyze the rover's current state and adjust the next sequence of instructions accordingly before uplinking commands to the rover \cite{mishkin2006prime}. Both the strategic and tactical cycle structures were inherited from the MER planning framework.

As mentioned above, the MSL mission introduced a `supratactical' cycle into the standard planning framework. This is a direct consequence of Curiosity's advanced and extensive scientific capabilities \cite{chattopadhyay2014mars}. On MSL, a single science campaign can last more than a week, due to both the large number of international researchers involved in the process and the complexity of the numerous sensors and analyzers on board the rover. This timeframe is too slow (fast) to be supported by the tactical (strategic) planners. The supratactical cycle therefore serves as an intermediary for multi-sol scientific campaigns, and acts as a bridge between the international scientific teams and the engineering personnel directly communicating with the rover.

\subsection{Curiosity's Navigation Modes}
\vspace*{1mm}

Curiosity was initially equipped with three primary navigation modes, each with a different level of autonomy \cite{grotzinger2012mars}. Blind-drive (the first mode) involves letting the tactical team assess the terrain around the rover and then prepare a sequence of instructions for the \emph{exact} manoeuvres to accomplish (e.g., drive 2 metres forward, turn 90 degrees right, etc.). As the rover executes these commands, it keeps track of its own motion using wheel odometry. Since this is a dead-reckoning technique (which causes the uncertainty of the state of the rover to increase with distance), blind-drive can only be used over short distances in most situations.

The second navigation mode, auto-navigation, employs a level of autonomy in order to reach a distant waypoint identified by the planning team. The rover frequently stops to survey the surrounding terrain using stereo vision, assess the traversability of the local region, and select a safe path to move closer to its goal. This process is accomplished using the Grid-based Estimation of Surface Traversability Applied to Local Terrain (GESTALT) algorithm, which is only able to detect and avoid geometric obstacles \cite{goldberg2002stereo}. In auto-navigation mode, Curiosity can also use visual odometry (VO) at a very low frequency (roughly every 10 metres) to verify the rover's egomotion (by performing ``slip checks'').

The final navigation mode is the slowest, but arguably the safest. In addition to navigating using obstacle detection and avoidance, Curiosity uses VO over short distances (1.5 metres or so) as an additional source of relative motion information. This is useful when fine positioning of the rover is required, or when driving on terrain with an elevated risk of  slip. Since 2012, other drive modes have been uploaded to the rover (such as visual target tracking) or derived from already-existing capabilities (e.g., combinations of blind-driving, VO, and geometric terrain assessment) \cite{maimone2016a}.

Curiosity inherited its VO-based capabilities for online slip detection and improved state estimation from the MER program \cite{maimone2007two}. Although relatively simple to measure once it is happening, slip is very hard to predict from vision data only. Mission planners have therefore largely relied upon empirical `slip versus slope' curves, for bedrock, cohesive soil, and loose sand, derived from terrestrial experiments in simulated Martian environments \cite{heverly2013traverse}.

\section{Long-Distance Navigation Challenges} 
\label{challenges}

As with any robotic exploration mission in an uncontrolled and partially unknown environment, a number of unexpected factors may affect mission execution or lead to outcomes different from those initially anticipated. In the case of MSL, two key parameters for evaluating mission performance are the number of soil samples successfully analyzed by the onboard laboratory and the total distance driven \cite{grotzinger2012mars}. Initially, mission planners anticipated that Curiosity should be capable of driving 18 kilometres and analyzing 11 samples within its warranty period (the first 687 sols of the mission). However, as of sol 1237, almost twice the mission's warranty period, the rover's onboard laboratory had just analyzed its 12th sample, and it had traversed less than 12.9 kilometres \cite{mslnaiflog, sutter2017evolved}. Although there is no doubt that the MSL mission has been highly successful, such a realization raises questions about what factors, exactly, may have contributed to slower overall navigation progress (besides those purely related to various science campaigns), and how their impact can be reduced in future missions. The following subsections provide an overview of contributing operational and Martian environmental factors.

\subsection{Operational Factors}

In 2016, a study of the various mission aspects influencing the productivity of the MSL operations group was conducted, with the goal of reducing the quantity of labor-intensive planning tasks and improving current and future mission performance levels \cite{gaines2016productivity}. A key element that stood out from the study is that, because Curiosity has very little computational power and limited general autonomy \cite{maimone2016a}, the commands uplinked to the rover on a regular basis must be detailed instructions rather than more generic goals. Such detailed command sequences necessitate an assumption of the rover's state and resources throughout the entire daily plan. Since such knowledge is uncertain (especially for long command sequences), highly conservative estimates are often made, resulting in a recurrent under-use of the rover's time and energy. On ``traverse'' sols (when driving is the main activity), these issues can be mitigated to some extent by leveraging Curiosity's navigation autonomy capabilities (which enable goal-driven driving). On long traverses, it remains the case that safety margins (for example attitude boundaries, maximal slip, or wheel current draw) tend to be conservative and can slow the mission down.

Another operational difficulty arises from the difference in the rotation periods of the Earth and Mars (a day on Mars is 37 minutes longer than a day on Earth). Because tactical planners normally operate during daytime hours on Earth, the continuously varying time difference between Pasadena (where the NASA Jet Propulsion Laboratory is located) and Gale Crater generates recurrent inefficient periods. For example, if the uplink window at the end of the day on Earth occurs near the end of the day on Mars, Curiosity will have to wait until the following sol to execute the requested tasks. This leads to a ``restricted sol,'' during which few or no activities can be scheduled (since the tactical planners have to wait for Curiosity to finish the assigned tasks and communicate with Earth).

\subsection{Environmental Factors}
\vspace*{1mm}

The highly heterogeneous Martian terrain may be the most significant element affecting the navigation abilities of current rovers. Although Curiosity was designed to handle sandy, hard, and rocky terrains better than the MERs, it was not intended to be driven over sharp embedded rocks (formed through wind erosion and called ``ventifacts''). Driving over such rocks early in the mission resulted in numerous punctures to Curiosity's wheels, which in turn has dramatically changed how the rover is driven on Mars \cite{lakdawalla2014curiosity}.

One resulting change involves driving more in sandy environments, to avoid concentrated loads on the wheels as much as possible---the general result is lower driving distances per sol. This is particularly the case during the traverse of megaripples, which are aeolian sand accumulations covered by a coarser sand layer \cite{arvidson2017mars}. The successful crossing of megaripples including Dingo Gap and Moon Light Valley, and the failures experienced in Hidden Valley, revealed that the traversability of such formations is influenced by both their geometry (shape, wavelength and amplitude) and their material properties. The exact material properties cannot be determined by the rover itself and are very difficult for mission planners to infer remotely.

Similar traversability issues were encountered several times by the MERs on sandy terrain. For example, Opportunity remained stuck in the Purgatory dunes for 38 sols and took six sols to leave the Jammerbugt ripple \cite{arvidson2011opportunity}. Orbital imagery was later used to identify additional fields of large ripples, requiring the rover to make several detours. Unforeseen terrain properties, in fact, brought Spirit's mission to an end: in April 2009,  the rover broke through a poorly cemented thin crust and embedded itself into unconsolidated soil \cite{arvidson2010spirit}, where it remained trapped.

Operating a solar powered-rover on the Red Planet adds another source of vulnerability. Energy availability became a major constraint when planning activities for the MERs \cite{bresina2005activity} after their missions extended beyond their warranty periods. Energy generation rates are heavily influenced by the opacity of the atmosphere, the amount of dust covering the solar panels, and the frequency of natural cleaning events caused by the wind \cite{landis2005exploring}. In addition, as the seasons change, the paths utilized by the MERs must be adjusted due to the change in the maximal elevation of the sun in the sky. The typically lower energy levels limit the activities that can be accomplished in a single sol, while also decreasing the amount of data that can be downlinked to Earth \cite{mishkin2006prime}.

\section{Recent Relevant Work} 
\label{work}

Based on the issues raised in the previous section, it is clear that increased situational awareness and effective energy modelling would allow for better predictions of the behaviour of a rover as it is navigating; in turn, a more productive use of battery resources would be possible. This section presents a selection of recent research that could potentially contribute to an increase of navigation autonomy under energy constraints.

\subsection{Increased Situational Awareness}

Passive exteroceptive sensors, such as cameras, perceive natural electromagnetic radiation from the environment. This type of sensing is used extensively by Curiosity whenever it arrives at a new site: imagery provides dense details about the spatial content around the rover. As such, efforts to extract useful information from these data involve vision-based terrain classification. At present, due to the low processing power on board Curiosity, dense image processing must happen on Earth.

One of the solutions to the accelerated wheel damage issue on MSL was the development of a risk-aware navigation planning tool to assist tactical operations and reduce risks associated with human errors \cite{ono2015risk}. The goal of this work was to easily distinguish safe and hazardous terrain (especially embedded sharp rocks), and plan safe paths based on the physical configuration of the surrounding ground and the detected local terrain types. A random forest-based algorithm was used to classify NAVCAM (navigation camera) images, categorizing each pixel as belonging to one of five terrain types. These terrain types were determined using a set of meaningful intensity and gradient-based features extracted from the spatial context around each pixel in each image. The random forest architecture was suitable for this task, primarily because of its speed of execution, inherent robustness to irrelevant or noisy data, and ability to capture nonlinear relationships between features (which are common in planetary environments). The classified image data were finally combined with more traditional attitude-related constraints in a random geometric graph framework, to find optimal, safe paths that considered the placement of the rover's wheels.

A more modern terrain classification approach was employed as part of the Mars 2020 landing site selection campaign. The Mars Reconnaissance Orbiter is equipped with the High Resolution Imaging Science Experiment (HiRISE) camera, which is able to capture images of the Martian surface with a resolution of 25 centimetres per pixel. Although downsampled or cropped HiRISE images are used on a regular basis by mission planning teams, it is impossible for humans to thoroughly inspect large areas (i.e., Mars 2020 landing site candidates) at the full resolution. The Soil Property and Object Classifier (SPOC) system, built on a convolutional neural network, was trained to assign one of 17 terrain classes to each HiRISE  pixel. SPOC was able to achieve high accuracy with only very sparse training labels supplied by humans experts \cite{rothrock2016spoc}.

A separate SPOC deep classifier with a similar architecture was trained with MSL NAVCAM data, and then used to annotate images taken on sols 0 to 938. The results were correlated with thousands of previously-recorded MSL slip events, for each terrain type, and compared to the empirical Earth-based slip versus slope curves used by tactical planners. As expected, the slip versus slope data retrieved from Martian traverses were slightly different from the Earth-calibrated models, except for the case of sand, where the differences were dramatic. Once again, this is mainly due to the failure of vision-based methods to fully characterize sandy soil.

In order to resolve the inability to predict slip on sandy terrain, a different exteroceptive sensing technique has recently been suggested: thermal measurements of the ground, from which thermal inertia can be derived. The thermal inertia of a sandy terrain describes the rate at which the terrain gains or loses heat relative to the surrounding environment. This property is strongly characterized by the physical characteristics of the sand, such as density, particle size distribution, cementation and others. Cunningham et al.\  \cite{cunningham2017improving} recently demonstrated that considering thermal inertia can increase slip prediction accuracy on sandy terrain. Although thermal inertia estimates from orbital data have already been used to assess general traversability, work in \cite{cunningham2017improving} represents the first use of in-situ measurements for this purpose. The intuition behind the correlation between thermal inertia and traversability is that, at Martian atmospheric pressure, the physical factors affecting thermal inertia are also key factors influencing the amount of slip experienced by a rover on granular terrain. The work in \cite{cunningham2017improving} presented a two-experts model approach (considering both thermal inertia and terrain slope) for in-situ thermal measurement (using Curiosity's ground temperature sensor) and orbital measurement (using the Mars Odyssey spacecraft's thermal imager). For each case, a threshold value separating low from high thermal inertia sand and a slip versus slope curve for each regime were learned. This model exhibited a lower error than the traditional single-expert model, which only considers terrain slope.

\subsection{Energy Models}

Existing wheel-soil interaction simulators such as the Adams-based Rover Terramechanics and Mobility Interaction Simulator (Artemis) can already provide an estimate of required wheel torques and velocity profiles using a virtual rover model on simulated terrains \cite{zhou2014simulations}. An estimate of energy consumption can easily be computed from these values. Such simulations are however computationally expensive and are not suitable for use in an optimization framework. More `convenient' energy models are required for long-distance navigation planning.

Sakayori et al.\ have suggested a deep learning-based approximation \cite{sakayori2017energy} to the terramechanics simulations for rovers driving on sandy inclines. The authors used the Wong-Reece wheel-soil interaction equations and the dynamic model of a 4-wheeled rover to generate a set of predetermined terrain and robot configurations and the corresponding theoretical energy consumption values. These data were used to train a feedforward neural network to output power consumption based on three input parameters: the rover's target velocity, its heading angle, and the slope angle of the local terrain. The high accuracy achieved by the network was heavily influenced by the assumption that the simulated sandy environment was uniform; however, this network concept could be readily extended to incorporate soil mechanical properties as inputs.

A similar idea for reducing computations related to energy consumption is presented in \cite{fallah2013energy}. Instead of direct calculation, a series of lookup tables store energy consumption values for different `bins' of slope angle and terrain type, obtained from dynamic and terramechanics simulations. The bin reference values are used in concert with solar energy generation predictions (dependent on the alignment of the rover's top plate normal with respect to the sun) to provide an energy profile for a given, discretized path. In this case, the optimality of a path depends on the net energy balance, path length, and a risk factor (related to the terrain configuration). Similar to \cite{sakayori2017energy}, this framework can also be applied to real-world platforms. 

A great deal of work to date has focused on producing energy models and maps from empirical data collected by proprioceptive sensors on board planetary rovers. Although such models are typically biased towards simple terrain types in the context of real missions because of safety concerns, they generally offer accurate terrain representations and mobility predictions. Research by Martin et al.\ \cite{martin2014long} has examined methods to generate energy-optimal paths by driving through an unknown environment several times and post-processing the gathered data. The energy cost of a path is represented as a function of parameters related to mobility (rover velocity, terrain configuration) and includes a constant energy sink defining the power needs of the internal rover components. Since power consumption data points are collected along discrete paths in a continuous space, gaussian process regression is used for interpolation purposes and to provide an understanding of where detail is lacking in the path map. This information is used incrementally during the exploration phase until optimal planning across the whole space is possible. The approach has been demonstrated (statistically) on simulated flat terrains and through simplified tests in natural environments \cite{martin2015long}.

Otsu et al.\ \cite{otsu2016energy} have also analyzed ways to exploit proprioceptive measurements, developing a self-supervised learning framework that uses vibration data to train a system to associate visual information with energy consumption. A SVM-based terrain classifier is first trained on features extracted from processed time-series acceleration signals (in a supervised manner) to identify different terrain types. Separately, linear regression (with two parameters only) is applied to fit empirical energy versus slope data for each terrain type. The linear relationship is based on several assumptions: the rover drives at constant velocity and the traversable slope is limited and deforms linearly. A vision-based classifier is then trained with the output of the vibration terrain classifier, to identify visible soil types. Combined with the corresponding terrain slope information, energy consumption can then be predicted.  This approach has been successfully tested using a real rover platform on three different soil types across multiple terrains with different slope angles.

\section{Discussion} 
\label{discussion}

The recent research presented in the last section may lead to new opportunities for sustainable long-distance navigation planning with solar-powered rovers such as the MSR fetch rover. In the past, detailed navigation planning at the strategic level has been very restricted due to the limited amount of information provided by orbital data. Now, with the ability of algorithms such as SPOC to provide richer information about the Martian terrain, more reliable planning is possible from orbital imagery. Identifying preliminary and then global energy-optimal paths could better inform tactical planners in the context of the MSR mission. Better integration of the strategic and tactical teams during long-distance optimal navigation planning would reduce reactivity and improve predictability.

An efficient use of the resources on board the MSR fetch rover will be vital to the success of the mission. To achieve this, more accurate empirical energy models for predicting energy consumption and generation will need to be employed. More efficient use of the rover's time will also be highly valuable---this may be accomplished by carrying out navigation planning for multiple sols in a row, lessening the impact of restricted sols, solar conjunctions, reduced planning efforts during weekends or holidays, and other inefficiencies. Multi-sol navigation with solar-powered rovers will require a dynamic activity planner to prioritize different tasks (driving along energy-optimal paths, stopping to replenish batteries, selecting a proper location to ``sleep'' for the night, etc.) and to maintain robustness against varying energy generation rates. Goal-driven planning and online task prioritization has recently been investigated in \cite{gaines2017expressing}.

It is also important to note that greater levels of autonomy, through improved online situational awareness (via exteroceptive sensing) and/or incremental energy model learning, may be possible over the next decade with the development of space-qualified Field Programmable Gate Arrays (FPGAs). These co-processors are expected to accelerate image processing through parallel computing on board the Mars 2020 rover \cite{seablom20162015}.

Lastly, our laboratory is currently developing methods to improve solar rover autonomy in Martian environments by incorporating empirical rover energy consumption information with orbital data, including thermal inertia maps, digital elevation models, and imagery pre-categorized with identified terrain classes. These techniques will be tested on real rover platforms and will be validated in several Martian analogue environments on Earth.

\section{Conclusion}

In summary, long-distance planetary navigation involving solar-powered rovers will be critically important in the next decade, especially for NASA's upcoming Mars Sample Return mission. This paper highlighted the operational and extraterrestrial environmental factors limiting the current mobility planning framework, and reviewed recent work tackling issues related to situational awareness and improved energy models.

A more detailed analysis of energy-efficient and long-distance navigation in a completely autonomous manner could extend the current survey. The present paper did not report on this specific research area, in order to keep the discussion focussed on existing operational processes and flight hardware limitations.

\subsection*{Acknowledgements}

This work was supported in part by the Natural Sciences and Engineering Research Council (NSERC) of Canada. The authors wish to thank Dr.\ Masahiro Ono and Dr.\ Kyohei Otsu from the NASA Jet Propulsion Laboratory for their invaluable advice regarding ongoing research, which greatly assisted in the writing of this paper.

\bibliographystyle{isairas_arxiv_v1}
\bibliography{isairas_arxiv_v1}

\end{document}